# Multi-context Attention Fusion Neural Network for Software Vulnerability Identification

Anshul Tanwar, Hariharan Manikandan, Krishna Sundaresan, Prasanna Ganesan,
Sathish Kumar Chandrasekaran, Sriram Ravi

*Cisco Systems*

## Abstract

Security issues in shipped code can lead to unforeseen device malfunction, system crashes or malicious exploitation by crackers, post-deployment. These vulnerabilities incur a cost of repair and foremost risk the credibility of the company. It is rewarding when these issues are detected and fixed well ahead of time, before release. Common Weakness Estimation (CWE) is a nomenclature describing general vulnerability patterns observed in C code. In this work, we propose a deep learning model that learns to detect some of the common categories of security vulnerabilities in source code efficiently. The AI architecture is an Attention Fusion model, that combines the effectiveness of recurrent, convolutional and self-attention networks towards decoding the vulnerability hotspots in code. Utilizing the code AST structure, our model builds an accurate understanding of code semantics with a lot less learnable parameters. Besides a novel way of efficiently detecting code vulnerability, an additional novelty in this model is to exactly point to the code sections, which were deemed vulnerable by the model. Thus helping a developer to quickly focus on the vulnerable code sections; and this becomes the "explainable" part of the vulnerability detection. The proposed AI achieves 98.40% F1-score on specific CWEs from the benchmarked NIST SARD dataset and compares well with state-of-the-art.

## 1. Introduction

Software Code vulnerabilities are the fundamental reason for cyber-attacks. It is an inevitable situation in software products and risks the credibility of the enterprise post-deployment. Even Google, Facebook, Cisco, Microsoft resolve more than a hundred security bugs in their products every year and it incurs a huge share of the company's yearly revenue. It carries a financial perspective because companies spend a large share of their resources, generating a patch and rolling out changes to kernel code. This is justified by the many vulnerabilities reported in the Common Vulnerabilities and Exposures (CVE) every year [1]. There are large-scale efforts both at the academic and the industrial front to tackle such vulnerabilities in code. Despite enormous research in this domain, there is still a sparsity of powerful automated methods that can accomplish can catch security bugs ahead of time. The Common Weakness Enumeration (CWE) provides various categories for the manifestation of the vulnerability in C code [2].

Since the vulnerabilities in code pose a big security risk it is important to detect them as early as possible. Vulnerability finding can be achieved by code static analysis, dynamic analysis and mixed analysis. Static analysis methods operate on static call graph of the code that is not compiled to run, while dynamic analysis examines program behaviour during run. There are several commercial and open-source dynamic analysis tools [3, 4] and studies [5-7] that detect security issues in code. Static analyzers [8-11] are the most popular methods for detecting bugs because they explore all the

- *Anshul Tanwar - PRINCIPAL ENGINEER* E-mail: *atanwar@cisco.com*.
- *Hariharan Manikandan - SOFTWARE ENGINEER* E-mail: *hmanikan@cisco.com*
- *Krishna Sundaresan - VP ENGINEERING* E-mail: *ksundar@cisco.com*
- *Prasanna Ganesan - DIRECTOR ENGINEERING* E-mail: *prasgane@cisco.com*
- *Sathish Kumar Chandrasekaran - TECHINCAL LEADER* E-mail: *sathicha@cisco.com*
- *Sriram Ravi – MANAGER ENGINEERING* E-mail: *srravi@cisco.com*

possibilities for program flow on the static call graph, but suffer from unhandled input issues. In enterprise production systems, mixed analysis is used: static analysis is first employed to detect and resolve most of the CWE related issues and standardize code styling, which is followed by thorough run-time testing that includes regressions and white-box testing.

Although static analyzers are widely used engineering systems for tackling security vulnerabilities, they emit large number of false positives. Recently, AI based code analysis have gained popularity for their ability to leverage all historic data for deriving key insights. Abstract Syntax Tree (AST), program dependency graphs and raw source code text are some of the common representations used to train AI for vulnerability classification.

In this work, the proposed AI, being the first of its kind, builds on top of features extracted from the code AST. The AST path contexts representation of the function code is modelled as a variable length sequence through the encoder-decoder orchestration. The encoder performs multi-type feature extraction and forms an attention fusion of these diverse features. The encoder learns comprehensive attention-based weighting over all the AST paths in a routine and assigns higher weights to those paths that decisively affect the chances of the code being vulnerable. The decoder employs multi-head attention as a feature selection mechanism to learn salient encoded features for target classification. Overall, the model is set to automatically learns characteristic embeddings for code which can serve as discriminatory features for tracing the vulnerability kind. The classification probability threshold can be optimally applied to emit class predictions.

## 2. Related Works

Emerging trends in NLP has also been used to apply standard language model architectures to address the code vulnerability. Machine learning for deriving robust source code feature representations is an active research area being worked by many organizations. For vulnerability classification, conventional approaches would pick any one of the neural architectures like Convolutional Neural Networks (CNN), Recurrent Neural Networks (RNN) or attention to solve this problem.

Some of the most popular source code-based static analysis methods for code vulnerability finding include, code similarity-based methods [12, 13] and pattern-based methods [14-16]. Several works have employed bytecode with sequence classification models (like LSTM), conversion of AST/Program Dependency Graphs/Call Graphs to vector representations and extended features like Common Vulnerability Scoring System (CVSS), CVE and text features. A common technique utilized in most of the research were temporal modelling using transformers, attention mechanism and embedding learning algorithms.

Similarity methods learn semantic code equivalence for detecting vulnerable codes. They are highly sensitive to similar code styles and are affected by code cloning. The pattern methods learn robust vulnerability patterns that characteristically identify CWEs in code. To build an AI that achieves security expert-level precision in finding specific CWEs in code, featurization is important. The feature extraction methods should be able to capture all of the syntactic and semantic properties of code that accurately expresses the vulnerability.

One of the earliest works in the field, VulDeePecker utilized code fragments as a representational model for vulnerability detection by Li et. al [13]. The researchers had used program backward slices that are executed when hitting an API/library function call to a program. The resultant program slices are stacked into code gadgets which are pieces of code that follow a semantic sequence between each other. These code pieces were fit on a Bi-directional Long Short Term Memory (LSTM) neural network. Fang et. al treated programmed codes as running texts and applied text featurization techniques for mapping text to features. The authors had used Continuous Bag of Words model for learning semantically similar embeddings for code that were classified on a Light Gradient Boosting Machine (GBM) model [17]. Ning Guo et. al used graph based static analysis methods for extracting code slices from the program bytecode [18]. A LSTM was employed to learn from these codes with full supervision. Zhen Li et. al studied the Syntax based and Semantic (SySe) based characteristics of a program to derive vector representations to these vulnerability candidates [19]. These elements/tokens are mapped to a symbolic representation that is converted to the vector form. A deep neural network is used to encode the vulnerability patterns and perform deep classification. Harer et. al utilized the control flow graph to derive blocks of code that follow sequentially during execution [20]. These token streams were passed through word2vec model and classified on a TextCNN in an end to end fashion. Xin Li et. al [21] presented an approach that uses the code slicing on the system dependency graph with links pointing to control and data flow dependencies to obtain security slices. The token embeddings in the sequence were further processed through a 1d CNN for learning security vulnerabilities. Lu et. al performed contrast feature extraction, that is transforming codes into Attribute Relation File Format (ARFF) [22]. These features are classified on a Logistic regression (LR), Gradient boosting Decision Tree (GBDT) against each of the CWE categories. Zhou et. al explored code property graphs that inherently captures control flow (CFG), data flow (DFG) and statement flow (AST) in code [23]. The vector representations for the graph node set were aggregated over all timesteps using 1D convolutional layers.

Though a large number of previous studies have significantly contributed to the research, the generalizability to real world, noise-robust learning and interpretability of AI are still actively researched areas. There is still ongoing work to improve the base learning paradigm, which can leverage the full potential of AI for detecting security threats from complex enterprise software.

## 3. Contributions

The main aspects of the proposed model are three-fold.
   a. To learn effective feature representations that can improve class distinguishability, a fusion of multi-contextual features is being proposed. The standout factor of the proposed sequence model lies in leveraging the best aspects from each component learner:
      - Cross-correlation between feature-sets from the self-attention module
      - Auto-regressive feature sampling from the Bi-LSTM and
      - Neighbourhood location context from the CNN.
      
      This is a complete paradigm shift over the approaches like the transformer encoding, which learns the same kind of attention features through different parameters. Further, the presented architecture also subsumes the foundational piece of transformer, i.e. self-attention onto its encoder. On top of that, it leaves a regularization effect over the feature extraction scheme.
   b. To exploit salient encoder features towards the classification, the proposed work includes an attention decoder that draws features of relevance and suppresses noise from the diversely attended context maps. Specifically, the decoder plays the vital role of integrating the variably-sized receptive field information and robustly mining data patterns (that characterize a vulnerability) out of the encoder. This significantly boosts the recall factor of the model.
   c. Another major impact of the presented model comes from the efficient computation. While state-of-the-art AI for AST-based code vulnerability detection bag over 6M learnable parameters, the presented model achieves on-par results in order of just 3M million parameters. Besides, it has shown the capability to adaptively learn rich localization details that intuitively explain the underlying mechanism through which it performs predictions.

## 4. Proposed Work

The proposed sequence model is orchestrated as an encoder-decoder architecture that forms an attention fusion over multi-contextual code features to predict vulnerability. The end-to-end architecture of the proposed model is presented in figure 1.

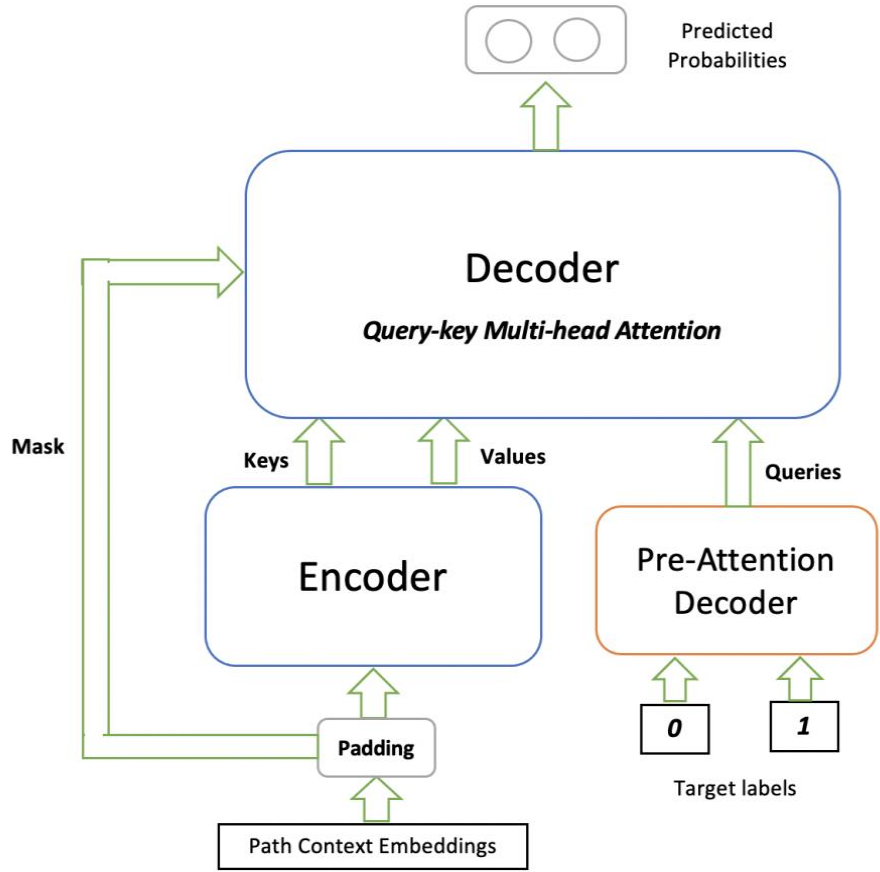

Fig. 1. Schematic Diagram of the proposed AI architecture

Features to train the model are derived as a set of concrete paths running between leaf nodes in the code Abstract Syntax Tree. The path nodes along with the terminal end nodes constitute a path context. By traversing the AST from left to right, such path contexts can be mined and framed into a sequence. A dataset of such AST path sequences extracted for the function codes is used to train the proposed model, which comprehensively learns an attention probability distribution over these AST features. The encoder applies multiple learning algorithms to generate a robust latent feature-set for the decoder.

The decoder uses pre-framed queries to identify relevant encoder features through a multi-head attention model. As a result, the generated hidden states at the decoder would have captured the weighing yielded to each part of the AST, which are exploited towards classification. The classifier is softmax activated and emits the probability of the sample being vulnerable.

### 4.1. Code Feature Generation

The code features as AST path contexts are derived as shown in figure 2. Path contexts in code are formed by traversing the path between any two AST leaf nodes and are represented in the form given by Eq. 1.

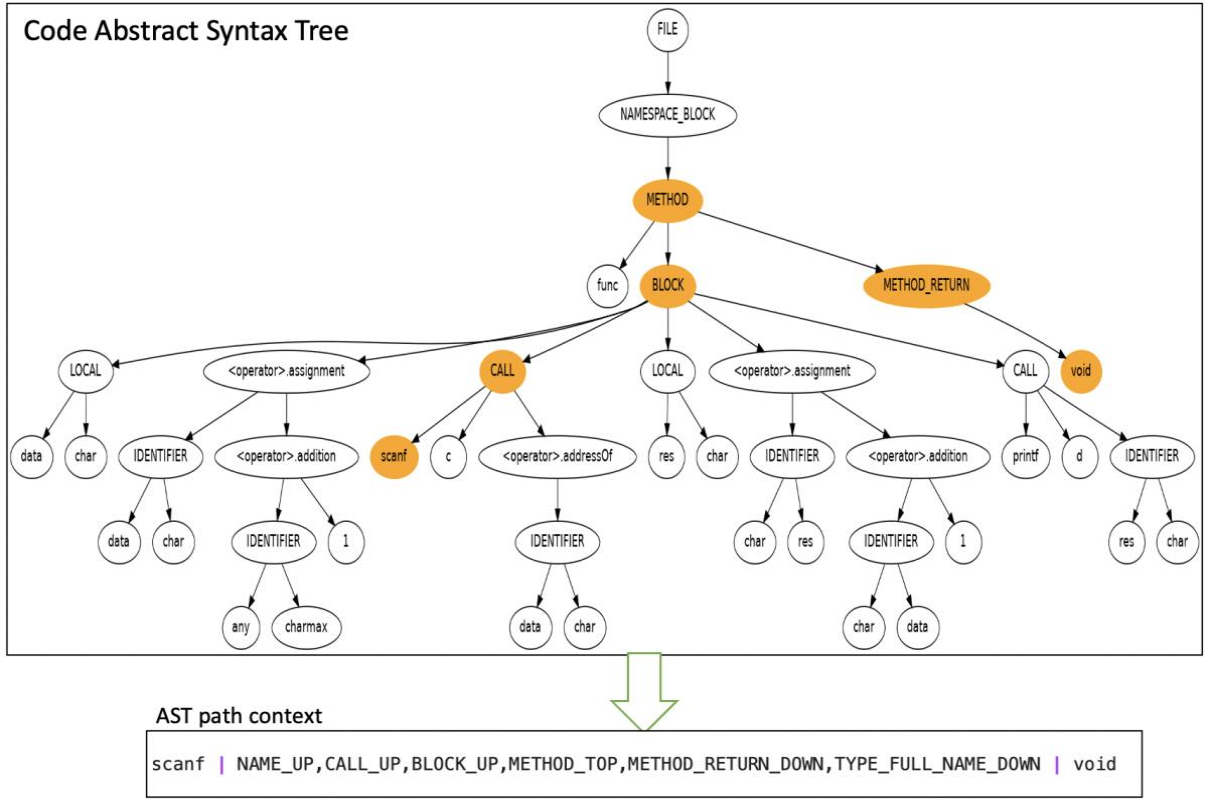

Fig. 2. Generating AST path contexts.

$$P: <x_s, p, x_t> \tag{1}$$

where, $x_s$ is the source node of the path context P

$x_t$ is the sink node of P

$p$ are the path nodes that occur in the path flowing from $x_s$ to $x_t$

Every path context is represented by the three components given in Eq. 1: source node, path nodes and sink node. In the example shown in Figure 2, the mappings are as follows:
- Source node, $x_s$: $scanf$
- Path nodes, path nodes $p$:  $Name \uparrow Call \uparrow Block \uparrow METHOD \downarrow Method\ Return \downarrow Type\ Full\ Name$
- Sink node, $x_t$: $void$

The arrows denote the direction of traversal with respect to the tree structure (up or down). The proposed encoder learns a neural path attention model for weighing all possible path contexts in code. It does that by first applying an embedding layer to these path contexts. Define two embedding mapping functions $\emptyset_N: node \rightarrow \mathbb{R}^D$ and $\emptyset_P: path \rightarrow \mathbb{R}^D$ that can be learned to generate D-dimensional vectors for source/sink nodes and the path. These embedding lookups are fit to the entire vocabulary of AST nodes and paths in the dataset. For path context $P$, the resulting vector representation $C$, is a concatenation of the node/path embeddings as shown in Eq. 2.

$$C = Concat <\emptyset_N(x_s), \emptyset_P(p), \emptyset_N(x_t)> \tag{2}$$

This joint context embedding, $C \in \mathbb{R}^{3D}$ denoting the path context forms the input seen by the AI for learning and prediction.

## 4.2. AST Path Sequence Encoder

The encoder is formed as a congregation of contextual features raised through different learning mechanisms, which are:
1) the Self-attention module,
2) Bi-directional LSTM and
3) 1D Convolutional network.

Figure 3 presents the architectural sketch of the encoder. The encoder aims to facilitate a learnable way for generating keys, i.e. the hidden states capable of achieving feature selection. Keys hold the compressed correlation information of the sequence that can be tuned to highlight features of interest for the decoder.

The advantages of pooling such diverse latent states from different models are three-fold:
1) It effectively regulates variably sized receptive field information at the fusion layer, which directly contributes to efficient feature mining.
2) Learning assortment of keys transfers knowledge of salient features for performing vulnerability class decoding of the inputs.
3) Aggregation of multiple contexts is useful in building an inferential model that can explain the choice of features mined towards identifying the vulnerability.

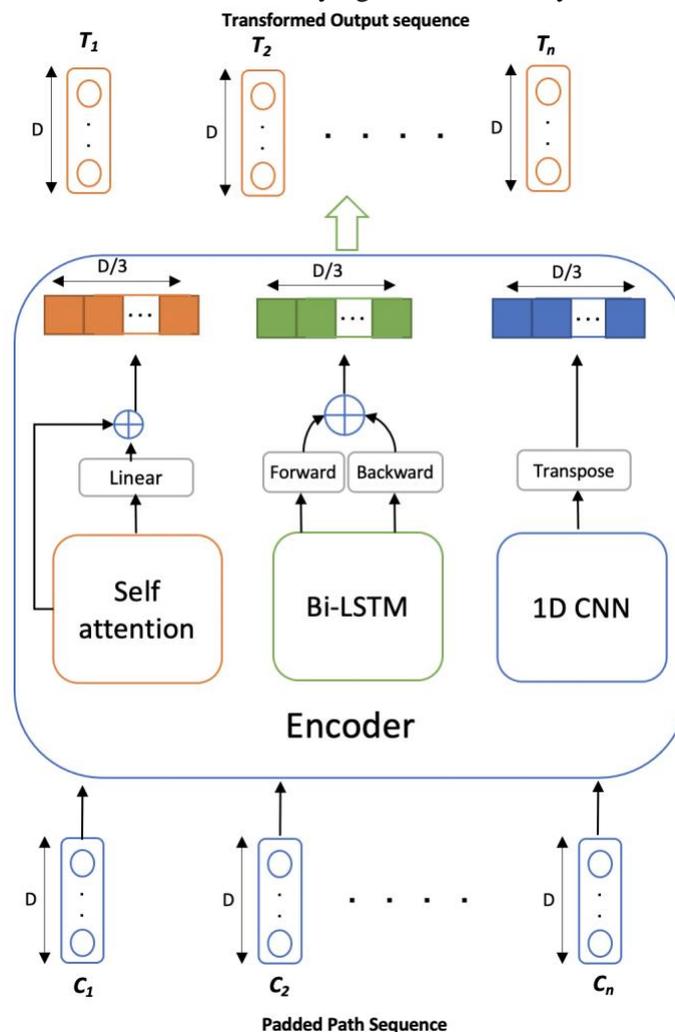

Fig. 3. The Path Attention encoder neural network

Let $f$ denote the context embedding sequence input for the encoder, where each time-step is D-dimensional. The encoder applies each of these learning paradigms, over the entire D-dimensions in the input, to result in an encoded sequence with constituent elements being one-third of D dimensions. Finally, the output hidden states (of D-dimensions) are formed as an assembly of these D/3-dimensional features pooled from each learning module in the array. This section shall further cover the three components of the encoder.

Firstly, the self-attention model performs correlations across states, $f_i$ in the $K$-length input sequence to yield the feature-set $A \in \mathbb{R}^{K \times D/3}$ as presented in Eq. 3. It is the generic QKV attention that applies a softmax-weighing over the correlations (inner product) between query-key and combines it against the intended values $V$.

$$A(Q, K, V) = softmax(QK^T)V \qquad (3)$$

where $Q = fW^Q$, $K = fW^K$, $V = fW^V$ are the self-attention queries, keys and values derived from $f$ respectively. The projection matrices $W^Q \in \mathbb{R}^{D \times D/3}$, $W^K \in \mathbb{R}^{D \times D/3}$, $W^V \in \mathbb{R}^{D \times D/3}$ are learnable parameters of the algorithm that generate attention features through linear transformations. The output $A$ is further distilled through a feed-forward layer and ReLU activation which are additively combined back with $A$ via a residual connection. The resulting features $S \in \mathbb{R}^{K \times D/3}$ for the $K$ keys in the sequence, are obtained as presented in Eq. 4.

$$S = A + ReLU(AW^l + b^l) \qquad (4)$$

where $W^l + b^l$ denote the weight and bias terms the feed-forward transformation.

Secondly, the bidirectional LSTM network wraps the context surrounding the sequence steps $t$ through cell states $c_t$. This recurrent model works by gating the cell states with input, forget, cell and output gates that help strip noise in forward prop and efficiently balance memory of cell states across the sequence. Bidirectionality adds to the robustness of the model by binding feature propagation in both directions and it provides dual set of hidden activation states corresponding to Left-to-Right or the opposite direction. The hidden state $h_{t-1}$ at time step $t - 1$ can be derived as a gated attention function over the input sequence. It is expressed through equations Eq. 5-6.

$$h_t = o_t \odot tanh(c_t) \qquad (5)$$

$$c_t = F_t \odot c_{t-1} + i_t \odot g_t \qquad (6)$$

where $\odot$ is the Hadamard product, $c_t$ is the cell state. $o_t, F_t, i_t, g_t$ are the output, cell, forget, input gates respectively. These are framed as a weighted linear combination on the input at step $t$. Let $h_t^L$ and $h_t^R$ denote the hidden states obtained for left-to-right and right-to-left LSTMs respectively. Then the joint feature activation $h_t$ at time step $t$ is computed as a sum of these forward and backward features. The resulting activations from the Bi-LSTM are the encodings $L \in \mathbb{R}^{K \times D/3}$.

The ternary set of keys $G \in \mathbb{R}^{K \times D/3}$, are extracted as 1-dimensional cross time-step convolutional features. The 1D CNN operation can be precisely described as performing a weighted cross-correlation (with weights $W^C \in \mathbb{R}^{D \times kernel\_size}$, bias $b^C \in \mathbb{R}$) over the input dimensions (or channels) $D_{in} = D$ to result in a single channelled output. Applying $D_{out} = D/3$ such weighted kernels, yields an output sequence with $D_{out}$ channels/dimensions. Each channel, 'o' in the output is a result of convolving with $W^o \in \mathbb{R}^{D \times kernel\_size}$ and bias $b^o$, for $o = 1$ to $D_{out}$.
Formulation of the 1D CNN hidden states output $C \in \mathbb{R}^{D_{out} \times K}$ is presented in Eq. 7.

$$C^o = b^o + \sum_{j=0}^{D_{in}-1} W_j^o * I_j \qquad (7)$$

where $*$ denotes the cross-correlation operator. $C^o$ is the encoded sequence for channel '$o$' in the output. $I \epsilon \mathbb{R}^{D_{in} \times K}$ is the input path sequence for the model. $j$ refers to the input channel in $I$ and $W^o$. Transpose of the resulting $C$ yields the keys $G \epsilon \mathbb{R}^{K \times D_{out}}$ where $D_{out} = D/3$.

The outputs from these diverse learners, $S, L, G$ are fused into a single feature map $E$ through linear concatenation along the feature axis and passed down the decoder.

### 4.3. AST Path Sequence Decoder

The attention decoder is constructed as a multi-head attention (MHA) model over the encoder keys $E \epsilon \mathbb{R}^{K \times D}$ and pre-framed queries $Q \epsilon \mathbb{R}^{2 \times D}$. The queries for the MHA are obtained from a pre-attention decoder as shown in figure 4. The query generator derives 2 distinct query vectors $Q = [q_0|q_1]$ corresponding to the vulnerable or non-vulnerable class (marked respectively as 1 or 0 in figure 4). An embedding layer generates these queries as dense vectors, followed by a few feed-forward layers.

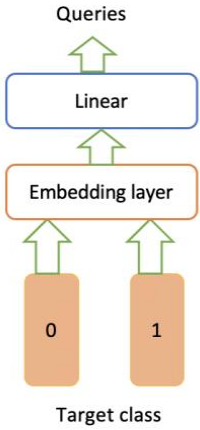

Fig. 4. Block diagram of the Pre-attention decoder

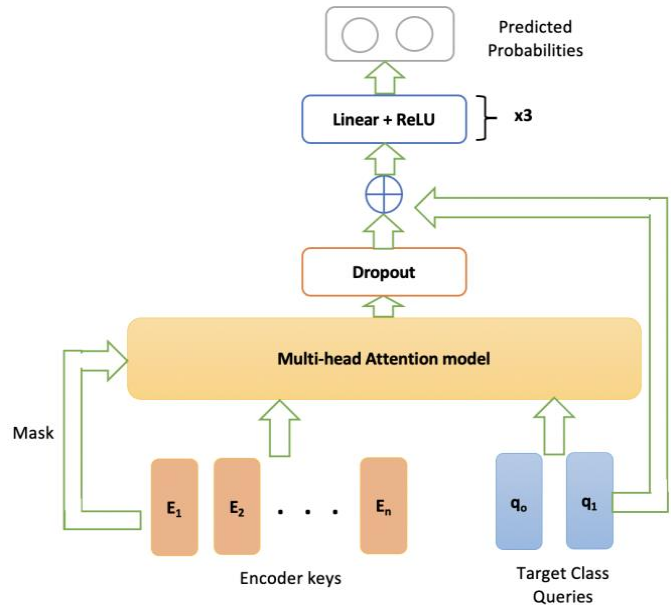

Fig. 5. Schematic diagram of the Decoder

The diagrammatic sketch of the decoder is presented in figure 5. The QKV attention principle in Eq. 3 forms the basis for MHA decoder. The MHA decoding works by spawning multiple attention heads ($= h$) that tend to the entire input dimensions D, but emit a fraction of D divided equally among the heads, i.e. $D/h$ dimensions at each head. Given queries Q and keys E, the MHA decoded sequence, expressed as a concatenation of outputs from $h$ heads is represented in Eq. 8.

$$MultiHead(Q, E) = Concat(head_1, \ldots, head_h)W^o \tag{8}$$

where $head_i = A(QW_i^Q, EW_i^K, EW_i^V)$ [refer Eq. 3]. The parameter matrices $W_i^Q \in \mathbb{R}^{D \times D/h}$, $W_i^K \in \mathbb{R}^{D \times D/h}, W_i^V \in \mathbb{R}^{D \times D/h}, W_i^V \in \mathbb{R}^{D \times D/h}$ are used to derive hidden states for the attention model. $W^o \in \mathbb{R}^{D \times D}$ linearly transforms the attended output.

The generated output from the MHA, say $T \in \mathbb{R}^{2 \times D}$ is subjected to dropout and added with the pre-attention queries through a residual link. Such residual learning helps prevent the vanishing gradients in complex network design. The feature maps are further propagated down multiple linear layers that sequentially funnel salient features for classification. The final feature tensor $\hat{y} \in \mathbb{R}^{2 \times 1}$ is softmax activated and optimized against cross-entropy loss for target label $y$, as given in Eq. 9.

$$L_{CE} = -\sum_{i=0,1} y_i \log \hat{y}_i \tag{9}$$

The probabilities $\hat{y}$ can be mapped to a class prediction by applying a suitable threshold value that is given by the model's ROC curve on the validation data.

## 5. Results of Model Training and Validation

The proposed AI was robustly evaluated on the benchmarked Juliet Test Suite that provides a dataset for 118 CWE vulnerabilities in software. Of these, experiments were performed on the data for the "Unchecked Return Value" vulnerability (CWE252), which had over 1700 samples of buggy and non-buggy codes. A stratified split of the dataset into train-dev sets in 4:1 ratio yields balanced-class samples across training and validation. The model was validated under a range of ML metrics including, precision, sensitivity, F1-score, specificity, area under ROC and cross-entropy loss. The results of model validation on these codes is presented in Table 1. Additionally, the model learning trend over the training epochs is tracked through figures 6-8. Figures 6 & 7 show the loss and accuracy curves recorded on the training-validation datasets, while figure 8 presents the Receiver-Operator characteristic (ROC) curve showing trade-off between True positive and False positive rate in shifting thresholds, on the validation dataset.

Table 1: Performance of the proposed model registered on the training and validation datasets for CWE252

| Phase | Precision | Recall | F1-score | Specificity | Area under curve | Mean entropy loss | No. of samples |
|---|---|---|---|---|---|---|---|
| Training | 99.86 | 99.71 | 99.78 | 99.87 | 99.99 | 0.0124 | 1466 |
| Validation | 98.92 | 97.87 | 98.40 | 99.39 | 99.95 | 0.0225 | 259 |

It can be observed that the AI achieves excellent train-test convergence. The effective gap between training and validation is constantly bridged throughout the learning phase. Though the loss decay seems staggering and non-monotonic, the model was able to retain the generalizability and lend well to the validation set. Similarly the accuracy curve is indicative of optimal model fitting. The ROC curve demonstrates the extent of separability of classes by placing a trade-off between TPR and FPR. From an area under ROC of 1.0 it is evident that the model shows superior class discriminability.

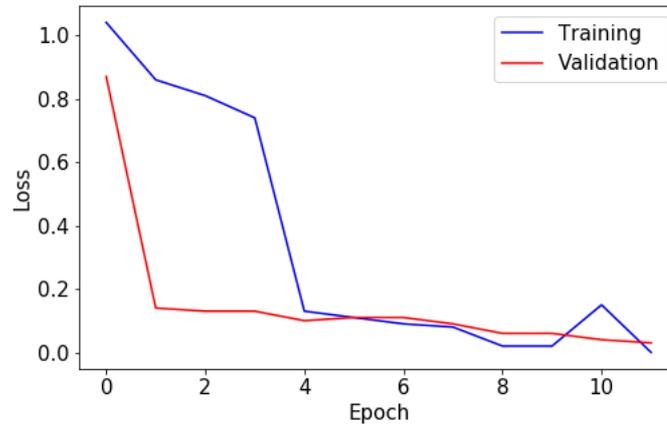

Fig. 6. Decay of the model training and validation loss with epochs

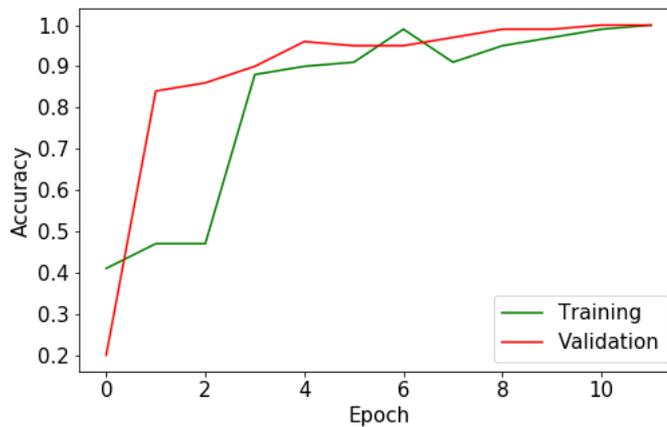

Fig. 7. Evolution of the prediction accuracy observed on the training and validation datasets with epochs.

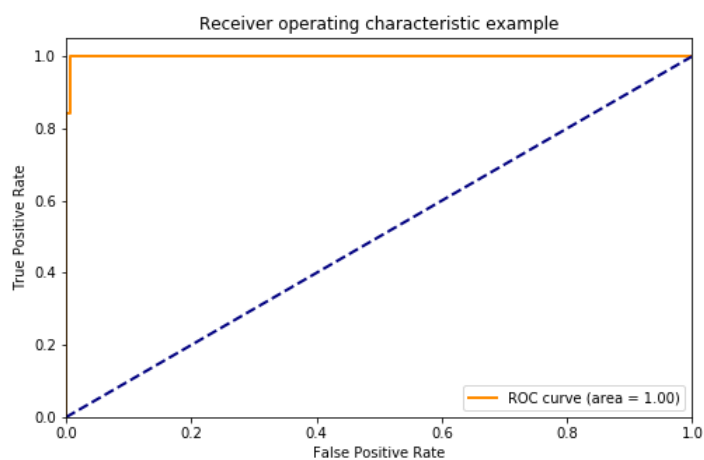

Fig. 8. Receiver-Operator characteristics registered on the validation dataset

With precision, recall, F1-scores of over 95% on the validation set, it is evident that the model has converged excellently. It is shown to eliminate false positives in addition to detecting a large share of real TP bugs. It has low miss rate and high detection accuracy, which directly compares with state-of-the-art. Specificity is the recall on the negative class and with 99% specificity, the model precisely identifies the non-buggy Juliet codes. Overall the model's performance on the Juliet dataset compares with state-of-the-art results. As an outcome of inspecting several patterns of vulnerability manifestation, the model was able to generalize well to unseen data in the validation set. Due to the attention orchestration, the model design is inherently explainable. As a next step, the model can be trained with large database of real-time CVE exposures from National Vulnerability Database (NVD). With more authentic vulnerability datasets, the predictive performance can be tuned by several folds.

### 5.1. Explainability of Predictions

Interpretation of the vulnerability prediction is critical for the next step of fixing the issue in code. Beyond just classification, the proposed AI is enabled to pinpoint the vulnerable lines. Given the prediction probabilities $\hat{y} \in \mathbb{R}^{2 \times 1}$ the exact set of input AST paths that have triggered the model's propensity towards a target class can be discerned through the attention head view technique.

The attention head view denotes the direct attention weights tended by the decoder on the encoder keys. Since the encoder hidden states are a result of fusing multi-context features, the decoder resolves the relevance of these feature-sets towards the target class query. The details of such a feature importance function learnt over the encoder keys are captured in the attention weights. For the target-class queries $Q = [q_0 \mid q_1] \in \mathbb{R}^{2 \times D}$, the aggregated attention head view of the decoder towards the encoder activations is presented Eqn. (10).

$$a_w = softmax(QE^T) \tag{10}$$

Here, $E$ is the linear concatenation of component keys hailing from the three learner modules in the encoder. $a_w$ expressed as an inner product between the keys $E$ and the queries $Q$, yields the affinities for each key against the target query. When the predicted $\hat{y}$ has a higher probability for the class $c$, the attention map corresponding to $a_w[c]$ will represent the contribution of each key in the input, as perceived by the neural network. These keys $E$ are directly a transformation of the respective AST path embeddings and therefore can be used to rank time steps in the sequence by the order of affinity.

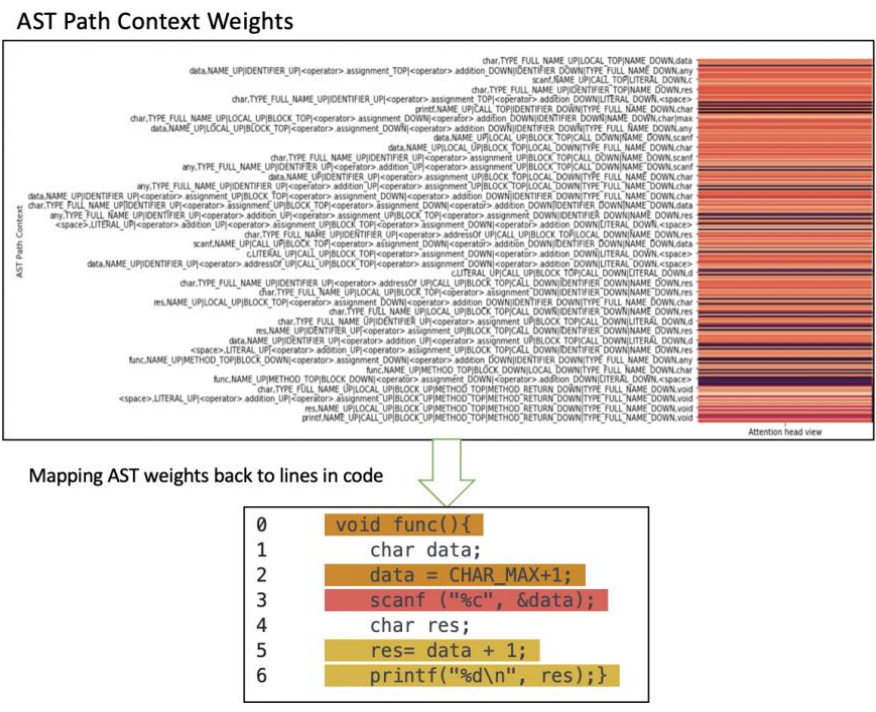

Fig. 9. Localization of CWE252 vulnerability to the lines in code

Figure 9 shows how the predicted attention scores can be explored to render line-level localization of the samples predicted as vulnerable. The example in figure 9 is the same AST from figure 2. All AST path contexts mined from the code sample were assigned attention weights. These weights attributed at path-level were further extended to AST nodes. Weightage for the component nodes comprising the path, were computed based on the degree of overlap of the nodes with different paths. The weightage attributed to these nodes are further resolved to lines in code. The resolution from AST back to code happens via creating a Directed Acyclic Graph (DAG) of the code and recursively mapping the AST child nodes along the breadth of the tree at each level to the DAG sections.

Codeline weightage (weighing given to each line of code) is calculated as weighted sum of their occurrences of AST path nodes across DAG sections in the code. In the example shown in figure 9, line no. 3 with the *scanf call* doesn't check the return value of the call and continues to handle the user entered variable throughout the scope of the program. A successful resolution of the vulnerability should have focussed on line 3 in arriving at the prediction. From the figure, it can be seen that each line of code has been highlighted with one of red, orange and yellow colors. The weight assignment to code is classified on three bands. The red lines have the highest probability of exposing the vulnerability, while the orange ones are statements that might possibly relate to lines marked in red. The yellow ones have low chance of qualifying as a vulnerability, but have non-zero attention weights, while the ones marked in white have absolutely zero attention weights. The AI model has shown the capability to precisely locate the hidden vulnerability to a large extent.

## 6. Conclusion

Resolving CWE issues in code well ahead of software release is critical for the software industry. It not only poses a serious security risk for the customer's data, but also impacts the quality of software development immensely. In this work, a multi-type attention fusion network was proposed to learn implicit weighing over path contexts in the AST. The underlying recurrent, convolutional and attention mechanisms enable the network to perform effective discriminatory feature mining that captures the semantic context of code, between different CWE classes. The AI is also built in an explainable manner, facilitating localizability of the vulnerability. By inspecting several samples, it was also validated that the attention model converged on the proper areas in the code context to focus on. The model can be further explored to understand what are the specific patterns observed by each component learner and to what aspects of the vulnerability are exposed by each of them. Further work in extending the model for several other CWE categories can be taken up. The attention localization can be enhanced with some level of supervision leveraging labelled data for the line of defects. The usability of the proposed architecture for different enterprise softwares and the suitability of AST, program dependency graphs as source code representations for building the model are future research directions.